\documentclass[runningheads]{llncs}
\usepackage[T1]{fontenc}
\usepackage{graphicx}
\usepackage{booktabs}
\usepackage[misc]{ifsym}
\newcommand{\corr}{(\Letter)}
\usepackage{mwe}
\usepackage{booktabs}
\usepackage{amsmath}
\usepackage{adjustbox} 
\usepackage{enumitem}
\usepackage{xcolor}
\usepackage{tcolorbox} 
\usepackage{multirow}
\usepackage{multicol}
\usepackage{colortbl}
\usepackage{CJKutf8}

\begin{document}

\title{A Unified Data Augmentation Framework for Low-Resource Multi-Domain Dialogue Generation}

\titlerunning{Low-Resource Multi-Domain Dialogue Generation}

\author{Yongkang Liu$^{\text *}$ \inst{1,2,4} \and
Ercong Nie$^{\text *}$ \inst{2,4} \and
Shi Feng\inst{1} \corr \and 
Zheng Hua\inst{2} \and
Zifeng Ding \inst{3,4} \and
Daling Wang \inst{1} \and
Yifei Zhang \inst{1} \and 
Hinrich Schütze \inst{2,4}}

\authorrunning{Y.K. Liu et al.}

\institute{Northeastern University, China
\and
Center for Information and Language Processing, LMU Munich
\and
Institute of Informatics, LMU Munich
\and
Munich Center for Machine Learning. LMU Munich \\
\email{misonsky@163.com, nie@cis.mu.de,zifeng.ding@campus.lmu.de}}

\maketitle

\def\thefootnote{*}\footnotetext{Equal contribution.}

\begin{abstract}
Current state-of-the-art dialogue systems heavily rely on extensive training datasets.
However, challenges arise in domains where domain-specific training datasets are insufficient or entirely absent.
To tackle this challenge, we propose a novel data \textbf{A}ugmentation framework for \textbf{M}ulti-\textbf{D}omain \textbf{D}ialogue \textbf{G}eneration, referred to as \textbf{AMD$^2$G}.
The AMD$^2$G framework consists of a data augmentation process and a two-stage training approach: domain-agnostic training and domain adaptation training.
We posit that domain corpora are a blend of domain-agnostic and domain-specific features, with certain representation patterns shared among diverse domains.
Domain-agnostic training aims to enable models to learn these common expressive patterns.
To construct domain-agnostic dialogue corpora, we employ a \textit{\textbf{de-domaining}} data processing technique used to remove domain-specific features. 
By mitigating the effects of domain-specific features, the model trained on the de-domained corpora can effectively learn common expression patterns in different domains.
Subsequently, we adapt the learned domain-agnostic features to the target domain through domain adaptation training.
We conduct experiments on Chinese dialogue datasets from five different domains and show that AMD$^2$G achieves superior performance compared to both direct training on the target domain corpus and collective training on all five domain corpora. 
Our work underscores AMD$^2$G as a viable alternative solution for low-resource multi-domain dialogue generation. Code and data associated with our work are available on an anonymous
GitHub repository\footnote{https://github.com/misonsky/Amdg}.

\end{abstract}

\section{Introduction}
The efficacy of established sequence-to-sequence methodologies in constructing dialogue systems has demonstrated remarkable success in previous research~\cite{serban2016building,liu2022pvgru,liu2022mulzdg,li2023enhancing}. More recently, the notable achievements of Large Language Models (LLMs), including Blender~\cite{roller2021recipes}, Meena~\cite{adiwardana2020towards}, ChatGPT~\cite{ouyang2022training}, and GPT-4~\cite{openai2023gpt4}, have prompted the research community to increasingly embrace generative models as the go-to approach.
However, training these models often requires huge corpora. Unfortunately, the availability of domain-specific corpora remains notably limited across many domains, including medicine, finance, and military, due to concerns on security, copyright, and other constraints~\cite{hathaliya2020exhaustive}.
On the other hand, while the current LLMs exhibit excellent comprehension and generation capabilities, they often suffer from hallucinations in tasks with obvious domain characteristics and strong factuality~\cite{bang2023multitask,liu2023evaluate,ji2023survey}, particularly in low- and zero-resource contexts.
Hence, devising strategies to leverage corpora from disparate domains to enhance the performance and accuracy of the target domain remains a relevant pursuit in the era of LLMs.

Existing methods for cross-domain dialogue generation can be categorized into three groups~\cite{qin2020dynamic}: i) Separate Pattern~\cite{wen2018sequence,qin2019entity,wu2021domain,nie2022cross,li2023adaptive}; ii) Mixed Pattern~\cite{madotto2018mem2seq,wu2019global,lin2021bitod,kim2023bidirectional,yang2023multi,ma2023domain}; and iii) Shared-Private Pattern~\cite{zhong2018global,chen2018multinomial,wu2019shared,wu2019transferable,bang2023task}. The Separate Pattern involves training the model separately for each domain, necessitating an adequate corpus for each domain. 
In contrast, the Mixed Pattern combines multi-domain datasets by prioritizing domain-agnostic features while disregarding domain-specific ones.
However, models based on the Mixed Pattern may struggle to capture features from low-resource domains compared to high-resource ones.
The Shared-Private Pattern extracts domain-agnostic and domain-specific features using shared and private modules, respectively, also relying on sufficient domain corpora.
While these methods have demonstrated promising results, they all presuppose the availability of ample corpus data.

\begin{figure}[t]
\centering
\includegraphics[width=0.70\linewidth]{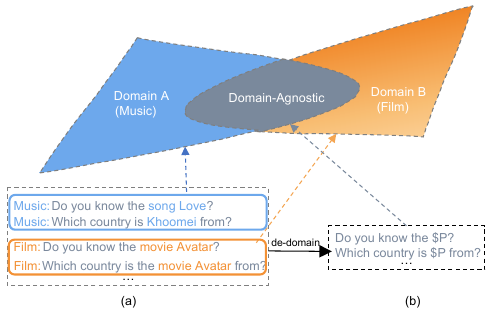}
\caption{Illustration of corpus composition in different domains. (a) represents domain-specific corpora, (b) stands for domain-independent corpora. The overlap of Domain A (blue) and Domain B (Orange) represents domain-agnostic data, while non-overlapping regions signify domain-specific data.}
\label{intro:example}
\end{figure}

In the context of dialogue generation within low-resource domains, the effective leverage of resources from other domains holds significant importance. As illustrated in Figure~\ref{intro:example}, the domain corpora encompass both domain-agnostic and domain-specific information. For instance, in the \textcolor{orange}{film} domain, the expression \textit{``Do you know the movie Avatar"} domain comprises domain-specific information \textit{``\textcolor{orange}{movie Avatar}''} alongside the domain-agnostic fragment \textit{``\textcolor{gray}{Do you know the ...}''}. Similarly, phrases like \textit{``\textcolor[RGB]{122,166,230}{song Love}"} and \textit{``\textcolor[RGB]{122,166,230}{Khoomei}"} carry domain-specific features of the \textcolor[RGB]{122,166,230}{music} field. Notably, disparate domains exhibit shared expression patterns within their corpora. Leveraging these shared features provides an opportunity to enhance the performance of dialogue generation in low-resource domains.

\begin{figure*}[t]
\centering
\includegraphics[width=\linewidth]{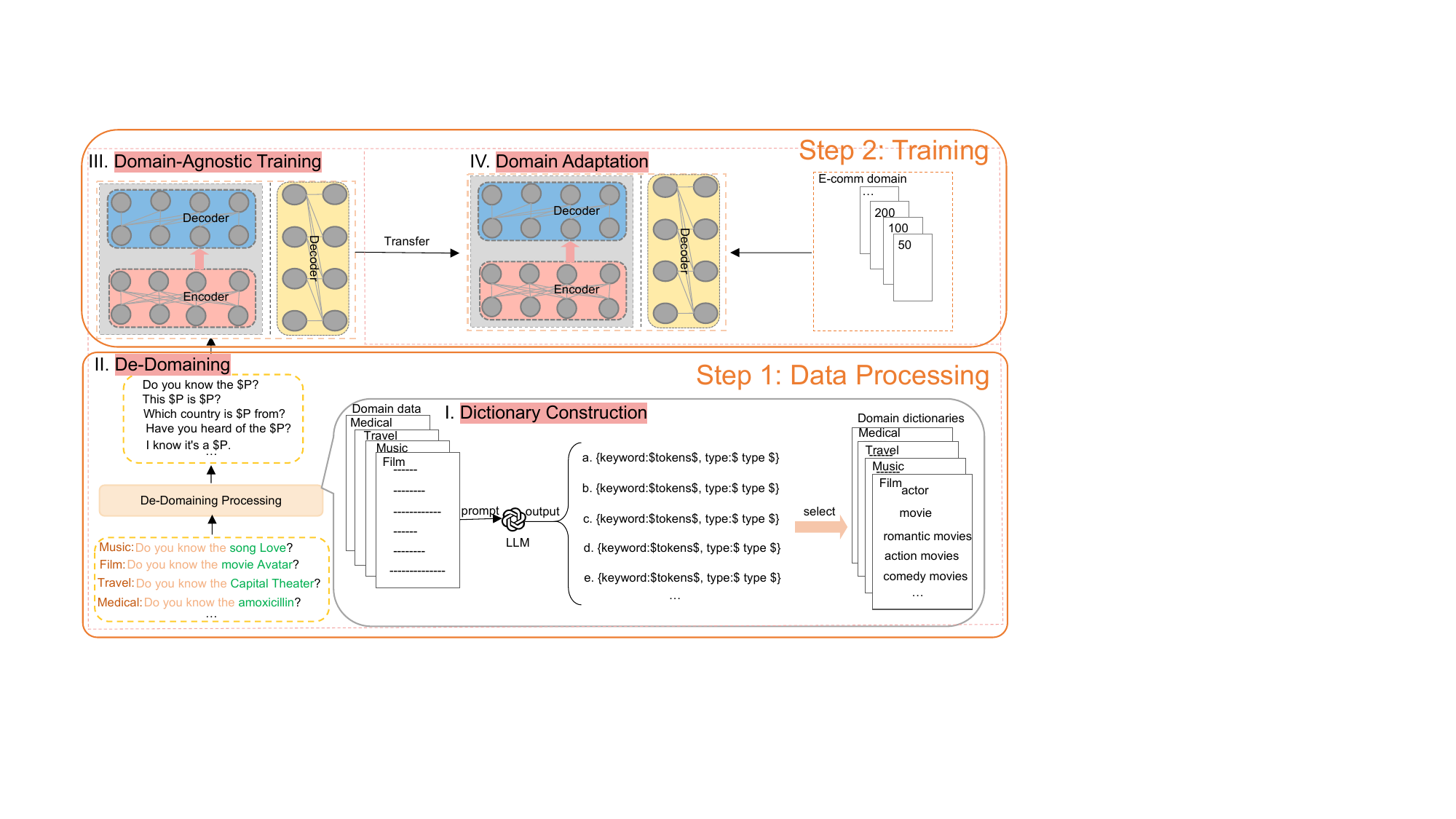}
\caption{Schematic diagram of \textbf{AMD$^2$G} framework. The target domain is E-Commerce and the domains used for de-domaining are Film, Music, Travel, and Medical. 
$\$P$ represents the placeholder. The method supports both \textbf{encoder-decoder} and \textbf{decoder-only} structures.}
\label{intro:framework}
\end{figure*}

Accordingly, we propose a simple yet effective data \textbf{A}ugmentation method for \textbf{M}ulti-\textbf{D}omain \textbf{D}ialogue \textbf{G}eneration, termed \textbf{AMD$^2$G}. As shown in Figure~\ref{intro:framework}, we initially build a domain dictionary for each domain automatically. Subsequently, the domain corpora undergo de-domaining through the usage of domain dictionaries. Specifically, placeholders replace domain-specific keywords in the corpus identified in the domain dictionary. 

The models are then fine-tuned on the combined de-domained corpora to learn common representation patterns across different domains. Finally, low-resource fine-tuning is conducted on the target domain dataset to acquire domain-specific features.
We conduct experiments with AMD$^2$G using Chinese conversation datasets from five distinct domains. This choice is motivated by the morphological simplicity of the Chinese language, characterized by few inflectional variations, aligning well with our de-domaining processing at the lexical level.
We compare our proposed method with two baselines: direct training on the target domain corpus and training on all five domain corpora collectively.
Our experimental findings demonstrate that the integration of AMD$^2$G consistently enhances model performance across all five domains.

In summary, our contributions are as follows:
\begin{itemize}[leftmargin=*]
    \item We introduce \textit{\textbf{de-domaining}} for multi-domain dialogue generation datasets, a data augmentation technique that effectively reduces the impact of domain-specific features by extracting shared representations across domains.
    \item We propose \textbf{AMD$^2$G}, a simple yet effective alternative framework for the multi-domain dialogue generation task in low-resource settings.
    \item We conduct experiments with AMD$^2$G across five domains, demonstrating its superiority over direct training in the target domain and joint training across all five domains.
\end{itemize}

\section{Methodology}
As depicted in Figure~\ref{intro:framework}, AMD$^2$G primarily comprises two steps: data processing and training. In data processing, the domain corpus is de-domained by using the constructed domain dictionaries. 
In the training step, domain-agnostic training first allows models to learn shared patterns among multiple domains, while the domain adaptation phase enables models to capture domain-specific features.

\subsection{Problem Formulation}
An instance for one domain can be represented as $(C,R)$, where $C$=$\{u_1,u_2,...,u_n\}$ with $n$ utterances representing the context of the dialogue. Here, $u_i$ represents the $i$-th utterance, and $R$ represents the corresponding response. 
Our goal is to build a corresponding dialogue generation system $P(R|C)$ using corpora from other domains under the low-resource condition in the target domain. 
Please refer to the experimental settings for detailed settings of low resources (i.e., Section~\ref{sec:setting}).

Note that AMD$^2$G can be applied to models of both encoder-decoder and decoder-only structures. For models with an encoder-decoder structure, the encoder is responsible for encoding the dialogue history $C$, and the decoder generates responses $R$ based on the encoded representation. For models with a decoder-only structure, we concatenate the dialogue history $C$ and responses $R$ into a consecutive sequence and perform sequence modeling by autoregression.

\subsection{De-Domaining Data Processing}
We observe that data from different domains exhibits shared representation patterns. Through de-domaining operations, corpora from different domains can be transformed into a unified space devoid of domain-specific features. This process mitigates the influence of domain-specific features, facilitating the learning of domain-independent features by models.

\subsubsection{De-Domaining}
Domain-specific corpora are de-domained based on the usage of domain dictionaries. Specifically, we replace all words or phrases present in the domain dictionary with designated placeholders, wherein all tokens within an involved phrase are substituted by a single placeholder. As a result of the de-domaining processing, the domain-specific data no longer retains its specific characteristics.

\subsubsection{Dictionary Construction} 
We combine LLM-extracted terms with existing term banks to construct high-quality domain dictionaries tailored to specific domains. Initially, we employ TechGPT~\cite{TechGPT}, a Chinese LLM, to extract domain entities as keywords. TechGPT has been enhanced for various information extraction tasks through the integration of domain knowledge graphs~\cite{ren2018techkg} facilitated by BELLE~\cite{BELLE}, which is specialized in extracting domain keywords. We apply the following prompt to TechGPT to extract entities from the domain corpora. \$Context\$ represents the context composed of dialogue history and response. \$Domain\$ represents the domain name.

\begin{tcolorbox}
    [colback=gray!20, colframe=gray!100, sharp corners, leftrule={3pt}, rightrule={0pt}, toprule={0pt}, bottomrule={0pt}, left={2pt}, right={2pt}, top={3pt}, bottom={3pt}]
{\small 
\textbf{Prompt: }\texttt{\begin{CJK}{UTF8}{gbsn}文本是\end{CJK}\$Context\$.\begin{CJK}{UTF8}{gbsn}领域是\end{CJK}\$Domain\$. \begin{CJK}{UTF8}{gbsn}请输出\end{CJK} \$Domain\$\begin{CJK}{UTF8}{gbsn}关键词. \end{CJK}}  \\
\textbf{Translation: }\texttt{The context is \$Context\$. The domain is \$Domain\$. Please output keywords related to \$Domain\$}.} \\
(\texttt{\$Domain\$} $\in$ \texttt{\{Film, Music, Travel, Medical, E-commerce\}})
\end{tcolorbox}

In addition to keywords extracted by the LLM, we utilize existing terminology banks offered by Chinese input method providers, including QQPinyin\footnote{\url{http://cdict.qq.pinyin.cn/v1}}, Baidu\footnote{\url{https://shurufa.baidu.com/dict}}, and SougouPinyin\footnote{\url{https://pinyin.sogou.com/dict/}}. We retrieve the terms specific to each domain from these sources and merge them into the respective domain's dictionary. Table~\ref{tab:dict} presents the statistic overview of the dictionary, including its size, coverage ratio concerning the training set, and the number of replaced tokens. Notably, the domain dictionary's coverage rate for the training corpus exceeds 95\%, with a significant reduction in domain-specific terms.

\begin{table}[ht]
\scriptsize
\centering
\caption{Statistic overview of domain dictionaries. \textit{\#Keyword} represents the number of keywords in the dictionary. \textit{\#Cov} represents the proportion of training set examples covered by the dictionary. \textit{\#RToken} represents the number of domain words removed from the training set.} 
\begin{adjustbox}{max width=0.60\columnwidth, center}
\begin{tabular}{cccc}
\toprule
Domain  & \#Keyword & \#Cov   & \#RToken \\
\midrule
Film    & 2463  & 100\%   & 38680     \\
Music   & 1427  & 100\%   & 32305     \\
Travel  & 1006  & 100\%   & 33820     \\
Medical & 18749 & 99.80\% & 37065     \\
E-comm  & 1384  & 95.10\% & 19602   \\
\bottomrule
\end{tabular}
\end{adjustbox}
\label{tab:dict}
\end{table}

\begin{table}[ht]
\scriptsize
\caption{Similarity scores between different domains. O2\textit{\$Domain\$} represents the average similarity score of other domains to \textit{\$Domain\$}. \textit{Uni}, \textit{Bi}, \textit{Tri}, and \textit{Quad} represent the recall rates of 1-gram, 2-gram, 3-gram, and 4-gram respectively. All values are magnified by a factor of 100.}
\begin{adjustbox}{max width=0.60\columnwidth, center}
\begin{tabular}{ccccc}
\toprule
Domain    & Uni    & Bi     & Tri    & Quad  \\
\midrule
O2Music   & 77.67~ & 32.41~ & 15.73~ & 7.80~ \\
O2Travel  & 67.01~ & 25.42~ & 12.30~ & 5.62~ \\
O2Film    & 63.91~ & 23.30~ & 11.13~ & 5.41~ \\
O2Ecomm   & 80.95~ & 28.80~ & 9.43~  & 2.26~ \\
O2Medical & 63.53~ & 16.15~ & 5.55~  & 1.86~ \\
\bottomrule
\end{tabular}
\end{adjustbox}
\label{tab:sim}
\end{table}

\subsection{Domain-Agnostic Training and Domain Adaptation}
We conduct the first stage of fine-tuning on a mixed domain-agnostic corpus excluding the target domain to learn domain-independent features. Models pay more attention to domain-independent features in the domain-agnostic training phase. The purpose of domain adaptation is to transfer domain-independent knowledge to the target domain, allowing models to learn domain-specific features. Subsequently, models are initialized with the weights from the domain-agnostic training stage and then fine-tuned on the low-resource target domain corpus.

\subsection{Domain Similarity}

We believe that the similarity between domains is an important factor for AMD$^2$G. To explore the impact of different domains, we propose a simple domain similarity evaluation method. Specifically, we employ n-gram recall between domains as the similarity metric. Given that the n-gram sets of domains A and B are $A_n$ and $B_n$ respectively, the similarity score of domain A relative to domain B is:
\begin{equation}
  \textbf{Similarity}_{A2B} = \frac{\sum A_n \cap B_n}{\sum B_n}
\end{equation}
The similarity of domain B to A $\textbf{Similarity}_{B2A}$ can be computed in the same way. $\textbf{Similarity}_{*2B}$ represents the degree of similarity between other domains and domain B. Theoretically, the greater the similarity between the mixed data and target domain, the greater the benefits of data augmentation to the target domain. Table~\ref{tab:sim} shows the average similarity scores between domains.
Note that all results are based on the training set after removing domain words. 
Expression paradigms usually consist of more than two words, so similarity scores based on 2-gram and above can better show the degree of similarity between domains. An elevated similarity score signifies a greater overlap of paradigms between the target domain and other domains.
\begin{table}[!t]
\scriptsize
\caption{Overview of results.
\textbf{Bold} indicates the best result. \textbf{target} represents the result of training on the corresponding domain training set. \textbf{mix} represents the result corresponding to the mixed training set of all domains.
\textbf{AMD$^2$G} represents the result based on the AMD$^2$G framework. \textbf{AVE} represents the average performance. \textbf{PPL} refers to perplexity.
}
\begin{adjustbox}{width=0.90\textwidth, center}
\begin{tabular}{cc|cccccccccc|cc}
\toprule
Model & Corpus  & BLEU-1 & BLEU-2 & BLEU-3 & BLEU-4 & Rouge-L & Dist-1 & Dist-2 & \multicolumn{3}{c|}{Embed A/E/G} &  AVE$\uparrow$    & PPL$\downarrow$     \\
\midrule
\multicolumn{14}{c}{Film}  \\
\midrule
\multirow{3}{*}{Transformer}  & target                  & 23.68~ & 13.66~ & 9.41~  & 6.65~  & 31.20~ & 17.11~ & 38.69~ & 74.36~ & 56.67~ & 84.43~ & 35.59~ & 6.7480~ \\
            & mix                     & 22.60~ & 12.72~ & 8.51~  & 6.07~  & 29.91~ & 20.16~ & 48.86~ & 77.47~ & 58.19~ & 86.69~ & 37.12~ & 6.4009~ \\
           & AMD\textsuperscript{2}G & 29.67~ & 16.96~ & 11.92~ & 8.70~  & 33.52~ & 22.92~ & 51.18~ & 78.64~ & 59.94~ & 86.87~ & \textbf{40.03}~ & \textbf{6.2826}~ \\
\hline
\multirow{3}{*}{CPT}     & target      &            26.18~ & 14.65~ & 10.05~ & 7.24~  & 31.32~ & 25.34~ & 56.69~ & 78.07~ & 58.81~ & 86.67~ & 39.50~ & 4.4767~ \\
            & mix                     & 26.15~ & 14.90~ & 9.86~  & 7.01~  & 32.14~ & 26.10~ & 58.39~ & 78.28~ & 59.38~ & 86.92~ & 39.91~ & \textbf{3.9037}~ \\
            & AMD\textsuperscript{2}G & 28.78~ & 16.92~ & 13.20~ & 8.31~  & 33.49~ & 25.04~ & 57.02~ & 80.91~ & 60.15~ & 87.36~ & \textbf{41.12}~ & 4.0225~ \\
\hline
\multirow{3}{*}{GPT-2}    & target                  & 8.92~  & 4.53~  & 2.70~  & 1.55~  & 21.11~ & 6.46~  & 18.47~ & 75.87~ & 57.68~ & 87.83~ & 28.51~ & 7.7622~ \\
            & mix                     & 10.74~ & 5.20~  & 3.05~  & 1.73~  & 24.67~ & 10.11~ & 25.49~ & 75.11~ & 56.74~ & 86.77~ & 29.96~ & 7.6871~ \\
            & AMD\textsuperscript{2}G & 13.89~ & 6.90~  & 3.80~  & 2.61~  & 23.42~ & 11.30~ & 32.54~ & 78.89~ & 59.23~ & 87.90~ & \textbf{32.05}~ & \textbf{7.6083}~ \\
\hline      
\multirow{3}{*}{BART}      & target                  & 26.52~ & 14.81~ & 10.17~ & 7.50~  & 31.25~ & 29.39~ & 60.04~ & 76.93~ & 58.41~ & 86.75~ & 40.18~ & 3.9784~ \\
            & mix                     & 25.64~ & 15.23~ & 11.06~ & 8.46~  & 32.73~ & 29.47~ & 61.73~ & 78.32~ & 59.93~ & 86.39~ & 40.90~ & 3.4973~ \\
            & AMD\textsuperscript{2}G &29.80~ & 16.80~ & 11.02~ & 8.42~  & 35.43~ & 30.91~ & 62.97~ & 78.36~ & 59.38~ & 86.99~ & \textbf{42.01}~ & \textbf{3.4282}~ \\
\midrule 
\multicolumn{14}{c}{Music}    \\
\midrule
\multirow{3}{*}{Transformer}  & target                  &32.49~ & 18.20~ & 12.53~ & 9.45~  & 34.04~ & 13.65~ & 30.98~ & 77.17~ & 61.17~ & 86.58~ & 37.63~ & 4.8624~ \\
            & mix                     & 36.37~ & 21.23~ & 13.98~ & 9.72~  & 39.19~ & 17.12~ & 39.06~ & 81.93~ & 65.84~ & 88.75~ & 41.32~ & 4.2834~ \\
            & AMD\textsuperscript{2}G & 39.66~ & 22.97~ & 14.41~ & 9.51~  & 40.84~ & 19.89~ & 40.77~ & 82.28~ & 66.63~ & 88.22~ & \textbf{42.52}~ & \textbf{4.1936}~ \\
\hline            
\multirow{3}{*}{CPT}     & target                  & 33.52~ & 19.00~ & 11.76~ & 7.67~  & 36.78~ & 18.01~ & 43.09~ & 81.77~ & 64.14~ & 89.76~ & 40.55~ & 3.6496~ \\
            & mix                     & 36.13~ & 20.71~ & 13.42~ & 9.03~  & 38.57~ & 20.97~ & 48.82~ & 81.95~ & 65.10~ & 88.98~ & 42.37~ & 3.3757~ \\
            & AMD\textsuperscript{2}G & 37.41~ & 21.24~ & 14.23~ & 10.05~ & 39.80~ & 21.50~ & 50.68~ & 82.63~ & 65.89~ & 89.64~ & \textbf{43.31}~ & \textbf{3.2297}~ \\
\hline            
\multirow{3}{*}{GPT-2}     & target & 27.13~ & 14.37~ & 9.40~  & 6.67~  & 32.74~ & 13.42~ & 31.68~ & 79.34~ & 62.54~ & 88.21~ & 36.55~ & 4.0748~ \\
            & mix                     & 32.83~ & 17.96~ & 11.81~ & 8.27~  & 34.10~ & 14.71~ & 34.45~ & 79.28~ & 62.99~ & 87.21~ & 38.36~ & 3.5407~ \\
            & AMD\textsuperscript{2}G & 35.98~ & 19.49~ & 13.83~ & 9.53~  & 35.00~ & 18.81~ & 44.34~ & 81.99~ & 63.31~ & 88.40~ & \textbf{41.07}~ & \textbf{3.3353}~ \\
\hline            
\multirow{3}{*}{BART}      & target                  & 34.97~ & 21.09~ & 14.46~ & 10.69~ & 40.84~ & 22.74~ & 47.80~ & 82.91~ & 66.93~ & 89.59~ & 43.20~ & 3.1210~ \\
            & mix                     & 36.02~ & 22.35~ & 15.88~ & 12.04~ & 41.94~ & 22.55~ & 49.43~ & 82.21~ & 67.23~ & 88.68~ & 43.83~ & 3.2139~ \\
            & AMD\textsuperscript{2}G & 38.25~ & 23.99~ & 15.81~ & 14.62~ & 43.46~ & 23.53~ & 53.89~ & 82.59~ & 68.09~ & 89.69~ & \textbf{45.39}~ & \textbf{3.0367}~ \\
\midrule       
\multicolumn{14}{c}{Travel} \\
\midrule 
\multirow{3}{*}{Transformer} & target                  & 34.63~ & 27.17~ & 22.57~ & 19.43~ & 47.26~ & 11.55~ & 25.24~ & 80.44~ & 70.83~ & 86.92~ & 42.60~ & 2.3491~ \\
            & mix                     & 36.03~ & 28.52~ & 23.65~ & 20.58~ & 46.85~ & 12.34~ & 31.90~ & 83.90~ & 72.15~ & 90.22~ & 44.61~ & 2.3350~ \\
            & AMD\textsuperscript{2}G & 36.30~ & 28.40~ & 24.63~ & 22.68~ & 46.96~ & 14.16~ & 33.41~ & 81.58~ & 69.39~ & 89.82~ & \textbf{44.73}~ & \textbf{2.2389}~ \\
\hline
\multirow{3}{*}{CPT}       & target                  & 26.80~ & 20.07~ & 15.67~ & 12.84~ & 47.25~ & 19.19~ & 43.12~ & 83.59~ & 72.76~ & 89.25~ & 43.06~ & 1.8730~ \\
            & mix                     & 35.17~ & 28.41~ & 24.14~ & 21.19~ & 51.19~ & 17.41~ & 40.83~ & 85.24~ & 74.29~ & 90.56~ & 46.84~ & 1.8273~ \\
            & AMD\textsuperscript{2}G & 36.67~ & 28.73~ & 24.41~ & 21.91~ & 51.93~ & 17.07~ & 39.85~ & 86.48~ & 74.96~ & 91.29~ & \textbf{47.33}~ & \textbf{1.8032}~ \\
\hline            
\multirow{3}{*}{GPT-2}    & target                  & 32.41~ & 24.07~ & 19.70~ & 16.59~ & 39.80~ & 11.50~ & 28.15~ & 79.60~ & 68.55~ & 87.45~ & 40.78~ & 3.6998~ \\
            & mix                     & 36.90~ & 28.48~ & 24.03~ & 21.52~ & 39.27~ & 14.57~ & 34.07~ & 80.81~ & 68.73~ & 88.66~ & 43.70~ & 3.1826~ \\
            & AMD\textsuperscript{2}G & 37.65~ & 30.09~ & 25.88~ & 23.18~ & 44.60~ & 12.07~ & 27.44~ & 83.11~ & 71.77~ & 89.41~ & \textbf{44.52}~ & \textbf{3.0171}~ \\
\hline            
\multirow{3}{*}{BART}        & target                  & 31.49~ & 23.14~ & 17.45~ & 13.64~ & 47.29~ & 15.10~ & 35.70~ & 84.23~ & 72.90~ & 90.81~ & 43.17~ & 1.6796~ \\
            & mix                     & 31.25~ & 24.27~ & 19.45~ & 16.27~ & 49.91~ & 20.88~ & 44.68~ & 85.06~ & 74.07~ & 90.80~ & 45.66~ & 1.6497~ \\
            & AMD\textsuperscript{2}G & 36.23~ & 28.93~ & 24.18~ & 20.72~ & 51.57~ & 19.85~ & 41.85~ & 84.75~ & 74.34~ & 91.25~ & \textbf{47.37}~ & \textbf{1.6307}~  \\
\midrule         
\multicolumn{14}{c}{E-Commerce}       \\
\midrule 
\multirow{3}{*}{Transformer} & target                  & 13.51~ & 7.81~  & 5.15~  & 3.55~  & 20.48~  & 5.42~  & 16.54~ & 62.11~      & 49.47~ & 68.15~ & 25.22~ & 2.3847~ \\
            & mix                     & 13.87~ & 8.00~  & 5.09~  & 3.51~  & 20.62~  & 9.11~  & 30.43~ & 62.79~      & 49.45~ & 69.53~ & 27.24~ & 3.2139~ \\
            & AMD\textsuperscript{2}G & 14.92~ & 9.71~  & 6.49~  & 5.97~  & 21.84~  & 9.37~  & 31.70~ & 64.67~      & 51.54~ & 69.17~ & 28.54~ & 2.3438~ \\
\hline            
\multirow{3}{*}{CPT}         & target                  & 13.25~ & 8.28~  & 5.70~  & 3.97~  & 23.07~  & 12.02~ & 36.63~ & 63.42~      & 51.04~ & 70.80~ & 28.82~ & 2.1533~ \\
            & mix                     & 15.55~ & 9.67~  & 6.91~  & 5.33~  & 22.29~  & 12.88~ & 40.92~ & 63.83~      & 50.41~ & 71.87~ & 29.97~ & 2.5117~ \\
            & AMD\textsuperscript{2}G & 16.54~ & 8.72~  & 5.66~  & 4.05~  & 24.17~  & 12.69~ & 41.93~ & 64.38~      & 50.97~ & 71.71~ & 30.08~ & 2.1168~ \\
\hline            
\multirow{3}{*}{GPT-2}       & target                  & 7.21~  & 3.69~  & 2.17~  & 1.37~  & 13.70~  & 3.01~  & 9.91~  & 60.58~      & 46.49~ & 71.09~ & 21.92~ & 2.2633~ \\
            & mix                     & 5.52~  & 2.64~  & 1.62~  & 1.16~  & 10.50~  & 3.60~  & 11.01~ & 58.52~      & 44.01~ & 69.00~ & 20.76~ & 2.2301~ \\
            & AMD\textsuperscript{2}G & 7.44~  & 3.13~  & 1.46~  & 0.76~  & 12.50~  & 5.36~  & 18.34~ & 60.49~      & 44.77~ & 71.60~ & 22.58~ & 2.2175~ \\
\hline            
\multirow{3}{*}{BART}        & target                  & 14.94~ & 9.04~  & 6.10~  & 4.18~  & 21.95~  & 13.23~ & 38.38~ & 62.88~      & 50.43~ & 70.25~ & 29.14~ & 2.0949~ \\
            & mix                     & 15.60~ & 10.56~ & 8.23~  & 6.99~  & 22.30~  & 13.78~ & 40.75~ & 62.39~      & 49.92~ & 71.05~ & 30.16~ & 1.9479~ \\
            & AMD\textsuperscript{2}G & 14.04~ & 9.25~  & 6.50~  & 4.70~  & 23.66~  & 18.61~ & 51.59~ & 62.97~      & 50.61~ & 70.91~ & 31.28~ & 1.8750~ \\
\midrule           
\multicolumn{14}{c}{Medical}   \\
\midrule
\multirow{3}{*}{Transformer} & target                  & 13.50~ & 8.94~  & 6.30~  & 3.13~  & 42.44~  & 1.95~  & 4.41~  & 80.71~      & 70.83~ & 82.43~ & 31.46~ & 1.9193~ \\
            & mix                     & 15.72~ & 9.93~  & 6.58~  & 3.27~  & 40.04~  & 6.93~  & 18.02~ & 79.17~      & 69.12~ & 81.27~ & 33.01~ & 2.4432~ \\
            & AMD\textsuperscript{2}G & 18.50~ & 11.33~ & 7.24~  & 3.98~  & 37.81~  & 9.39~  & 25.10~ & 78.39~      & 67.17~ & 81.52~ & 34.04~ & 1.8951~ \\
\hline            
\multirow{3}{*}{CPT}         & target                  & 14.03~ & 9.29~  & 6.46~  & 2.49~  & 40.85~  & 13.68~ & 29.23~ & 78.31~      & 68.46~ & 81.43~ & 34.42~ & 1.9884~ \\
            & mix                     & 19.88~ & 12.50~ & 8.32~  & 4.62~  & 39.37~  & 13.75~ & 36.04~ & 78.69~      & 67.54~ & 82.21~ & 36.29~ & 1.6892~ \\
            & AMD\textsuperscript{2}G & 21.90~ & 13.69~ & 8.48~  & 4.37~  & 39.99~  & 14.34~ & 37.87~ & 79.40~      & 68.18~ & 82.78~ & 37.10~ & 1.6861~ \\
\hline            
\multirow{3}{*}{GPT-2}       & target                  & 19.13~ & 11.12~ & 6.82~  & 3.08~  & 35.33~  & 4.09~  & 10.46~ & 80.67~      & 70.22~ & 83.00~ & 32.39~ & 6.0602~ \\
            & mix                     & 23.70~ & 14.16~ & 8.90~  & 4.33~  & 36.42~  & 4.53~  & 12.10~ & 81.14~      & 70.92~ & 83.32~ & 33.95~ & 5.7444~ \\
            & AMD\textsuperscript{2}G & 23.94~ & 15.02~ & 8.30~  & 4.17~  & 36.67~  & 4.78~  & 14.60~ & 82.45~      & 71.45~ & 83.23~ & 34.46~ & 5.7559~ \\
\hline            
\multirow{3}{*}{BART}        & target                  & 13.05~ & 8.84~  & 6.35~  & 3.01~  & 42.69~  & 8.22~  & 15.05~ & 79.97~      & 70.55~ & 82.21~ & 32.99~ & 1.9705~ \\
            & mix                     & 16.67~ & 11.12~ & 7.79~  & 4.43~  & 42.77~  & 11.72~ & 27.45~ & 79.74~      & 69.65~ & 82.79~ & 35.41~ & 1.9147~ \\
            & AMD\textsuperscript{2}G & 22.97~ & 12.29~ & 7.94~  & 4.56~  & 41.22~  & 12.29~ & 29.75~ & 80.97~      & 68.91~ & 83.04~ & 36.39~ & 1.7770~ \\
\bottomrule             
\end{tabular}
\end{adjustbox}
\label{tab:com1}
\end{table}

\section{Experiments}

\subsection{Datasets}
In this paper, we experiment on Chinese dialogue generation datasets from five domains (i.e., Film, Music, Travel, Medical and E-commerce). Film, Music, and Travel are all from KdConv~\cite{zhou2020kdconv} datasets.  
KdConv is a Chinese multi-domain conversation dataset comprising 4.5K conversations from three domains: Film, Music, and Travel. It contains 86K utterances with an average turn number of 19.0. These conversations feature in-depth discussions and natural transitions between multiple topics. The \textbf{Film}, \textbf{Music}, and \textbf{Travel} domains contain 1,500 training sets, 150 validation sets, and 150 test sets respectively.
\textbf{MedDG} is a large-scale Chinese medical dialogue dataset, which contains 14,864 training sets, 2,000 validation sets, and 1,000 test sets respectively~\cite{liu2022meddg}. 
\textbf{E-commerce} is a large-scale e-commerce conversation dialogue dataset, containing 500,000 positive training examples, 1,000 validation sets and 1,000 test sets respectively~\cite{zhang2018modeling}. Similarly, we randomly selected 2,000 examples as the training set.
In order to meet the low resource setting, we extract 2000 conversations as the training set for \textbf{MedDG} and \textbf{E-commerce}. It is worth noting that we do not use the knowledge base.

\subsection{Models}
To evaluate the effectiveness and robustness of our proposed method, we apply \textbf{AMD$^2$G} to different types of models, including encoder-decoder and decoder-only structures. For the encoder-decoder structure models, we employ the basic \textbf{Transformer}~\cite{vaswani2017attention} structure as well as Chinese pre-trained language models such as \textbf{CPT}~\cite{shao2021cpt} and the Chinese version of \textbf{BART}\footnote{\url{https://huggingface.co/fnlp/bart-base-chinese}}~\cite{lewis2020bart}. The decoder-only structure model used in our experiment is the Chinese version of the pre-trained model \textbf{GPT-2}\footnote{\url{https://huggingface.co/uer/gpt2-chinese-cluecorpussmall}}~\cite{radford2019language}.

\subsection{Baselines}
For each model, we first compare AMD$^2$G with training only on the original domain training set and training on the mixed training set of all domains.
Besides, we compared AMD$^2$G with two other multi-domain transfer methods: \textbf{TS-NET}~\cite{peng2019teacher} and \textbf{DA-NET}~\cite{ma2023domain}. TS-NET uses a teacher-student network mechanism to transfer knowledge from other domains to the target domain, while DA-NET uses domain attention to realize knowledge transfer. We apply different methods to two types of models. One is the pre-trained Chinese model, \textbf{BART}, and the other is a language model that adopts the architecture of gated recurrent neural networks (GRU), a variant of RNN~\cite{chung2014empirical}.

\subsection{Implementation Details}
\label{sec:setting}
We implement our model and baselines using the Huggingface library
and train baselines on a server with RTX A6000 GPU (48G). We consider at most 10 turns of dialogue context and 50 words for each utterance. The batch size is 32, the minimum decoding length is set to 10, the maximum decoding length is set to 128, the warmup steps are 1,000, and the initial learning rate is 5e-5. We use the AdamW~\cite{loshchilov2017decoupled} optimizer to update model parameters. 
The beam size and length penalty coefficient are set to 6 and 1.0 respectively. The random seed of the sampled data is set to 12345. The random seed of the training process is set to 12345. We use GLoVe~\cite{pennington2014glove} to train 300-dimensional word vectors based on the training set for evaluation. The values of hyperparameters described above are all fixed using the validation set. We explore more low-resource scenarios including the ratio of [5\%,10\%,20\%,30\%,40\%] of the target corpus.
\subsection{Evaluation metrics}
We follow the previous studies~\cite{liu2022mulzdg,li2017adversarial,lin2022modeling,xu2018diversity,liu2022pvgru}, using both automatic and human evaluations to assess the performance of models.
Automatic evaluations include BLEU~\cite{yang2018adaptations}, ROUGE-L~\cite{lin2004rouge}, Dist~\cite{li2016diversity}, and Embedded metrics. Perplexity is also measured as an additional metric.
For human evaluation, we request annotators to score the generated responses with respect to three aspects: fluency, diversity, and relevance.
Each dimension is divided into three levels: 0, 1, and 2. In terms of fluency, 0 means not fluency, 1 means average fluency, and 2 means very fluency. Other evaluation dimensions are similar to fluency. After collecting the assessments from annotators, the final score is the average of all samples. Note that we use an improved version of BLEU that is more in line with human evaluation, and the calculated score will be lower than the original BLEU~\cite{papineni2002bleu}.

\section{Results and Analysis}

\begin{table}[ht]
\scriptsize
\caption{The performance of \textbf{AMD$^2$G} compared with other baselines in film, travel, and e-commerce domains.}
\begin{adjustbox}{width=\textwidth, center}
\begin{tabular}{cccccccccccc|cc}
\toprule
Model   & Method  & BLEU-1 & BLEU-2 & BLEU-3 & BLEU-4 & Rouge-L & Dist-1 & Dist-2 & \multicolumn{3}{c|}{Embed A/E/G} & AVE$\uparrow$   & PPL$\downarrow$      \\
\midrule
\multicolumn{14}{c}{Film}    \\
\midrule
\multirow{2}{*}{BART}       & TS-NET                  & 27.75  & 14.43  & 9.76   & 6.55   & 33.94   & 28.45  & 61.44  & 78.22       & 58.42 & 85.77 & 40.47 & 5.6721   \\
           & AMD\textsuperscript{2}G & 29.80  & 16.80  & 11.02  & 8.42   & 35.43   & 30.91  & 62.97  & 78.36       & 59.38 & 86.99 & \textbf{42.01} & \textbf{3.4282}~  \\
\hline  
\multirow{2}{*}{GRU}        & DA-NET                  & 15.55  & 11.01  & 7.38   & 6.33   & 22.44   & 11.07  & 23.08  & 67.43       & 54.76 & 83.55 & 30.26 & 25.6935~ \\
           & AMD\textsuperscript{2}G & 16.70  & 13.52  & 8.43   & 8.01   & 25.38   & 12.61  & 25.78  & 69.35       & 56.05 & 84.42 & \textbf{32.02} & \textbf{21.3866}~ \\
\midrule  
\multicolumn{14}{c}{Travel}    \\
\midrule  
\multirow{2}{*}{BART}       & TS-NET                  & 37.22  & 20.99  & 15.22  & 12.68  & 40.87   & 22.45  & 51.09  & 80.67       & 66.53 & 87.02 & 43.47 & 5.0060~  \\
           & AMD\textsuperscript{2}G & 38.25  & 23.99  & 15.81  & 14.62  & 43.46   & 23.53  & 53.89  & 82.59       & 68.09 & 89.69 & \textbf{45.39} & \textbf{3.0367}~  \\
\hline             
\multirow{2}{*}{GRU}        & DA-NET                  & 17.86  & 14.55  & 10.22  & 8.55   & 27.69   & 14.58  & 27.82  & 71.84       & 56.62 & 85.33 & 33.51 & 12.6600~ \\
           & AMD\textsuperscript{2}G & 19.22  & 15.62  & 12.64  & 10.33  & 27.71   & 16.20  & 28.44  & 73.05       & 57.22 & 85.44 & \textbf{34.59} & \textbf{10.3700}~ \\
\midrule  
\multicolumn{14}{c}{E-Commerce}    \\
\midrule  
\multirow{2}{*}{BART}       & TS-NET                  & 13.69  & 10.04  & 6.72   & 4.55   & 21.79   & 16.93  & 29.02  & 60.44       & 49.79 & 69.94 & 28.29 & 1.9044~  \\
           & AMD\textsuperscript{2}G & 14.04  & 9.25   & 6.50   & 4.70   & 23.66   & 18.61  & 51.59  & 62.97       & 50.61 & 70.91 & \textbf{31.28} & \textbf{1.8750}~  \\
\hline             
\multirow{2}{*}{GRU}        & DA-NET                  & 9.33   & 5.66   & 3.78   & 2.28   & 16.64   & 4.57   & 15.77  & 60.54       & 48.72 & 66.12 & 23.34 & 7.7743~  \\
           & AMD\textsuperscript{2}G & 11.62  & 6.44   & 4.44   & 3.22   & 20.09   & 5.67   & 17.66  & 60.62       & 49.90 & 67.03 & \textbf{24.67} & \textbf{7.1090}~ \\
\bottomrule  
\end{tabular}
\end{adjustbox}

\label{tab:baseline}
\end{table}

\begin{figure}[!t]
\centering 
\includegraphics[width=\linewidth]{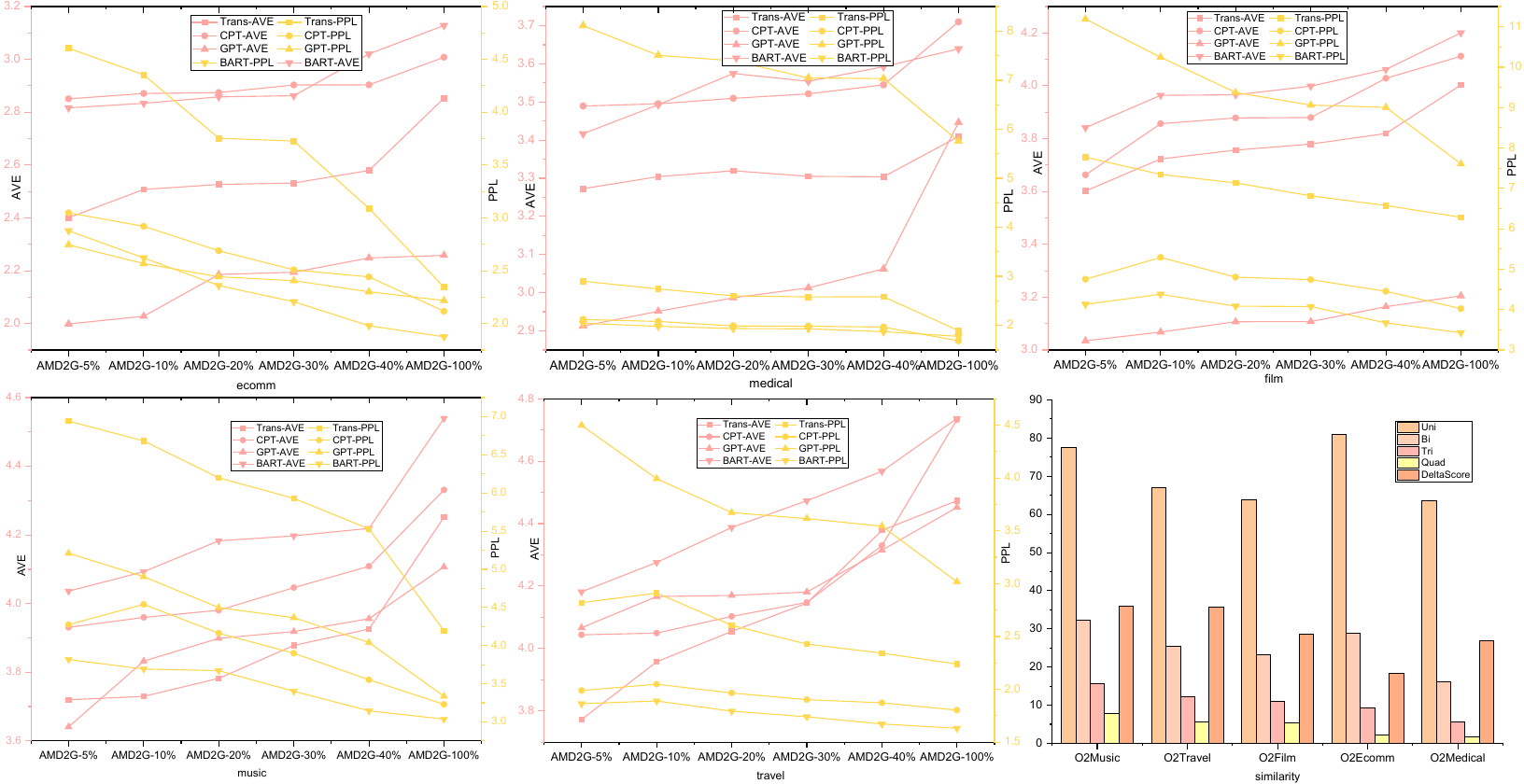}
\caption{The first 5 pictures show the trend of average performance and the trend of PPL as the training data changes in five domains. The last one is n-gram similarity score (i.e., Uni, Bi, Tri and Quad) and average performance gain trend (i.e., DeltaScore) of models based on \textbf{AMD$^2$G} compared to direct training on target domain corpus. To highlight the trend, we multiply the DeltaScore value by 1000.}
\label{img:figures}
\end{figure}

\subsection{Overall Results}
Table~\ref{tab:com1} reports the experimental results of AMD$^2$G in five domains. Compared with training directly on the target domain training set, AMD$^2$G demonstrates absolute advantages in five domains. Specifically, the average performance of the four models has improved by 1.85\% in the e-commerce domain, 2.69\% in the medical domain, 2.86\% in the film domain, 3.59\% in the music domain, and 3.58\% in the travel domain, compared with training directly in the target domain. Figure \ref{img:figures} shows that the performance of some models using 30\% target domain training corpus has achieved competitive performance based on AMD$^2$G framework compared to training on the target domain training set. When using 40\% target domain training corpus, the performance based on AMD$^2$G is equivalent to training directly on the target domain training set. These results fully demonstrate the effectiveness of AMD$^2$G.

According to Table~\ref{tab:sim}, even though the two domains are quite different, they still share some expression paradigms, which accounts for why the AMD$^2$G framework is effective. The first stage of the AMD$^2$G framework allows the model to learn shared paradigms between different domains and the second stage allows the model to learn domain-specific features, which can maximize the learning of domain-agnostic features and reduce the mutual influence between domain features. Compared with the performance of training when mixing all domain corpora, the performance of models based on AMD$^2$G exhibits absolute advantages, which confirms that features between domains can interfere with each other and has a negative impact on model performance. We will discuss in detail the impact of domain similarity on model performance in the next section.
Table~\ref{tab:baseline} reports the performance comparison of AMD$^2$G and other baselines. We can observe that AMD$^2$G has certain performance advantages compared to TS-NET and DA-NET. TS-NET uses a distillation mechanism 
to transfer domain knowledge, which relies on a large amount of target domain data. Low resource settings will severely impact distillation results. DA-NET utilizes a dynamic attention mechanism for multi-domain feature fusion. While it preserves domain-specific knowledge, it tends to overlook domain-independent features, thereby hindering its ability to effectively utilize features from other domains. The AMD$^2$G adopts a two-stage strategy
to retain domain-agnostic and domain-specific features, and domain-independent features can be adapted to the target domain and achieve data enhancement effects.

\subsection{Impact of Domain Similarity}

Domain similarity is a key impact factor for model performance. We perform further experiments to analyze the impact of similarity between domains on model performance. The last one in Figure~\ref{img:figures} shows the distribution of domain similarity based on n-gram and the average performance gain of models based on AMD$^2$G.
We find that the similarity based on 1-gram does not accurately reflect the similarity relationship between domains because it is greatly affected by the unified placeholder. In fact, scores based on more than 2-gram can better reflect the similar relationship between domains, because the common expression paradigm is based on more than two words. According to the 4-gram similarity score, the data enhancement effect based on the AMD$^2$G is more obvious for domains with high similarity scores to a certain extent. 
The more similar the domains are, the more helpful the knowledge provided by other domains will be to the target domain. 
An exception is the medical domain. Although the similarity score based on 4-gram is low, models based on the AMD$^2$G
framework have achieved a certain degree of gain. The domain characteristics of the medical domain are relatively obvious. After domain-free operations, the alignment of expression paradigms in different domains can be achieved to a certain extent. This can be derived from the similarity scores based on n-grams. This is why the AMD$^2$G framework can work in the medical domain.

\subsection{Impact of Dataset Size}

In order to explore the impact of data set size, we conduct experiments on 5\%, 10\%, 20\%, 30\%, 40\%, and 100\% of the training set, respectively. Note that 100\% of the training set refers to the total amount of data under low resources.
Figure~\ref{img:figures} reports the changing trends of model performance with data set size in five domains. We can observe that the average performance (i.e., AVE) basically increases gradually as the data size increases, and the PPL decreases as the data set size increases. Models are more sensitive to the size of the data set under low resource conditions. Even a small increase in the data set will make the performance of models increase. The increase in the data set will allow models to learn more different examples, making models cover more test examples.

\subsection{Human Evaluation Results}
\begin{table}[ht]
\scriptsize
\caption{The results of the human evaluation in e-commerce, film, and travel domains.}
\begin{adjustbox}{width=\textwidth, center}
\begin{tabular}{ccccccccccccc}
\toprule
\multirow{2}{*}{Model} & \multirow{2}{*}{Corpus}                  & \multicolumn{3}{c}{E-Commerce}  &  \multicolumn{3}{c}{Film}  & \multicolumn{3}{c}{Travel}           \\
\cline{3-11}
      &                         & Fluency & Relevance & Diversity  & Fluency & Relevance & Diversity  & Fluency & Relevance & Diversity \\
\midrule
\multirow{3}{*}{BART} & target                  & 0.377 & 0.833 & 0.062 & 0.579 & 0.875 & 0.083 & 1.034 & 0.920 & 0.116 \\
      & mix                     & 0.464 & 1.025 & 0.085 & 0.662 & 0.965 & 0.098 & 1.142 & 0.946 & 0.104 \\
      & AMD\textsuperscript{2}G & \textbf{0.522} & \textbf{1.150} & \textbf{0.113 }& \textbf{0.784} & \textbf{1.110} & \textbf{0.115} & \textbf{1.206} & \textbf{1.012} & \textbf{0.133} \\
\hline
\multirow{3}{*}{GPT-2} & target                  & 0.311 & 0.753 & 0.065 & 0.466 & 0.782 & 0.076 & 0.972 & 0.938 & 0.096 \\
      & mix                     & 0.472 & 0.975 & 0.086 & 0.673 & 0.950 & 0.120 & 1.133 & 0.955 & 0.110 \\
      & AMD\textsuperscript{2}G & \textbf{0.514} & \textbf{1.040} & \textbf{0.082} & \textbf{0.762} & \textbf{1.132} & \textbf{0.112} & \textbf{1.174} & \textbf{1.102} & \textbf{0.128}  \\
\bottomrule

\end{tabular}
\end{adjustbox}
\label{tab:human}
\end{table}

\begin{table}[!ht]
\scriptsize
\caption{Comparison with the human evaluation of baselines in e-commerce, film, and travel domains.}
\begin{adjustbox}{width=.8\textwidth, center}
\begin{tabular}{ccccccccccc}
\toprule
\multirow{2}{*}{Model} & \multirow{2}{*}{Method}                  & \multicolumn{3}{c}{E-Commerce}  &  \multicolumn{3}{c}{Film}  & \multicolumn{3}{c}{Travel}           \\
\cline{3-11}
      &                         & Fluency & Relevance & Diversity & Fluency & Relevance & Diversity & Fluency & Relevance & Diversity \\
\midrule
\multirow{3}{*}{BART}  & TS-NET                  & 0.456~  & 0.814~    & 0.065~    & 0.677~  & 1.067~    & 0.108~    & 1.110~  & 0.864~    & 0.105~    \\
      & AMD\textsuperscript{2}G & \textbf{0.522}~  & \textbf{1.150}~    & \textbf{0.113}~    & \textbf{0.784}~  & \textbf{1.110}~    & \textbf{0.115}~    & \textbf{1.206}~  & \textbf{1.012}~    & \textbf{0.133}~    \\
\hline
\multirow{3}{*}{GRU}   & DA-NET                  & 0.237~  & 0.755~    & 0.052~    & 0.508~  & 0.833~    & \textbf{0.117}~    & 0.456~  & 0.867~    & \textbf{0.088}~    \\
      & AMD\textsuperscript{2}G & \textbf{0.307}~  & \textbf{0.774}~    & \textbf{0.069}~    & \textbf{0.553}~  & \textbf{0.912}~    & 0.099~    & \textbf{0.542}~  & \textbf{0.955}~    & 0.086~    \\   
\bottomrule     
\end{tabular}
\end{adjustbox}

\label{tab:human_baseline}
\end{table}

We conduct the human evaluation of BART and GPT-2 on three domains to further confirm the effectiveness of AMD$^2$G.
To evaluate the consistency of the results assessed by annotators, we employ Pearson’s correlation coefficient~\cite{sedgwick2012pearson}. This coefficient is 0.25 on diversity, 0.68 on relevance, and 0.77 on fluency, with \textit{p} < 0.0001 and below 0.001, which demonstrates high
correlation and agreement. The results of the human evaluation are shown in Table~\ref{tab:human}. Compared to training directly in the target domain and on mixed corpora, models based on the AMD$^2$G enjoy a significant advantage in relevance and diversity. Specifically, models based on AMD$^2$G enjoy an average advantage of 20.40\% in fluency, 24.1\% in relevance, and 3.1\% in diversity in three domains compared with models training directly in the target domain. Compared with models training on mixed corpora, models based on AMD$^2$G enjoy an average advantage of 6.9\% in fluency, 10.2\% in relevance, and 1.3\% in diversity in three domains. The experimental results show the effectiveness of the AMD$^2$G framework. AMD$^2$G can effectively reduce the mutual interference of domain features and strengthen the learning of domain-agnostic features. Table~\ref{tab:human_baseline} reports the performance comparison of AMD$^2$G and other baselines. The human results show that the AMD$^2$G framework still outperforms the TS-NET and DA-NET. Specifically, the model's average performance, based on AMD$^2$G, is 9.8\% higher than TS-NET and 4.3\% higher than DA-NET.

\section{Related Work}
Dialogue systems are used as intelligent agents in various domains due to their ability to generate fluent and natural responses. Cross-domain learning refers to the technology of transferring knowledge from other domains to the target domain. The initial approach is to blend all domain corpus to learn domain-independent features. One of the drawbacks of this approach is the lack of domain-specific knowledge. Wu et al.~\cite{wu2019global}
propose to use a global-to-local pointer mechanism search technique for external knowledge to enhance the domain-specific knowledge awareness of models. He et al.~\cite{he2020fg2seq} employs pairwise similarity to distill contextually unrelated KB records to improve the quality of domain knowledge. Xie et al.~\cite{xie2022unifiedskg} integrates domain-specific knowledge in the form of text-to-text format based on T5~\cite{raffel2020exploring}.
Ma et al.~\cite{ma2023domain} propose a domain attention module with distributional signatures of dialogue corpus to capture domain-specific knowledge.   
These methods may also lead to a long-tail distribution of domain data, making models trained severely biased under low-resource conditions. Another line of research is to train separate models for each domain focusing on domain-specific features. Madotto et al.~\cite{madotto2018mem2seq} learns domain-specific features by combining external knowledge through a memory network~\cite{sukhbaatar2015end}. ~\cite{qin2019entity} proposes to employ two-step retrieval and attention mechanisms to improve the quality of domain-specific features. Wu et al.~\cite{wu2021domain} proposes to continue pre-training on domain corpus to adapt the language model to a specific domain.  The shared-private framework, which combines the advantages of the above two, is a better choice. Zhong et al.~\cite{zhong2018global} uses global modules to share parameters and local modules to learn domain-specific features. Wu et al.~\cite{wu2019shared} allows domain-specific features to interact through shared-private mechanism. Bang et al.~\cite{bang2023task} learns task-related features by adding adapters for each task.

\section{Conclusion}
We propose a simple and effective data augmentation framework AMD$^2$G for multi-domain low-resource dialogue generation. The domain characteristics of the corpus can be removed through domain dictionaries constructed by LLMs. Models trained on domain-independent corpus can reduce the interference of different domain features when models learn domain-independent features. Domain adaptation training can adapt the learned domain-independent features to the target domain.
Experiments on four models in five domains demonstrate the effectiveness of the AMD$^2$G framework. Compared with other baselines, the AMD$^2$G framework has obvious advantages.
AMD$^2$G provides an alternative solution for low-resource multi-domain dialogue generation.

\section*{Acknowledgement}
We would like to thank reviewers for their constructive comments. The project is supported by
the National Natural Science Foundation of China (62172086, 62272092) and DFG (grant SCHU 2246/14-1). The project is also supported by China Scholarship Council.

%
%
%
\bibliographystyle{splncs04}
\bibliography{mybibliography}
\end{document}